# Style transfer-based image synthesis as an efficient regularization technique in deep learning

Agnieszka Mikołajczyk  
Department of Electrical Engineering, Control Systems and Informatics, Gdańsk University of Technology, Poland  
agnieszka.mikolajczyk@ pg.edu.pl

Michał Grochowski  
Department of Electrical Engineering, Control Systems and Informatics, Gdańsk University of Technology, Poland  
michal.grochowski@pg.edu.pl

*Abstract*— These days deep learning is the fastest-growing area in the field of Machine Learning. Convolutional Neural Networks are currently the main tool used for the image analysis and classification purposes. Although great achievements and perspectives, deep neural networks and accompanying learning algorithms have some relevant challenges to tackle. In this paper, we have focused on the most frequently mentioned problem in the field of machine learning, that is relatively poor generalization abilities. Partial remedies for this are regularization techniques e.g. dropout, batch normalization, weight decay, transfer learning, early stopping and data augmentation. In this paper we have focused on data augmentation. We propose to use a method based on a neural style transfer, which allows to generate new unlabeled images of high perceptual quality that combine the content of a base image with the appearance of another one. In a proposed approach, the newly created images are described with pseudo-labels, and then used as a training dataset. Real, labeled images are divided into the validation and test set. We validated proposed method on a challenging skin lesion classification case study. Four representative neural architectures are examined. Obtained results show the strong potential of the proposed approach.

*Keywords—deep neural networks, regularization, neural style transfer, data augmentation, decision support system, diagnosis, skin lesions*

I. INTRODUCTION

Currently, systems based on deep neural networks (DNN) can reach near human-level performance in many types of tasks, such as classification, segmentation, natural language processing, image generation, signal processing and many more. Despite great achievements and bright perspectives, huge deep neural models with many layers and millions of parameters, still does not always generalize well enough even though they show a small difference between the training and test performance [1]. One way of making models generalize better is regularization, which currently is one of the most important techniques in the field of deep learning. Regularization is defined as any supplementary technique that aims at making the model generalize better, for example, produces better results on the test set [2]. A common regularizations strategies are: to inject random noise into various parts of the model (dropout [3]), to limit the growth of the weights (weight decay [4]), to select the right moment to stop the optimization procedure (early stopping [5]), to select the initial model weights (transfer learning [6]). Moreover, popular approach is to use dedicated optimizers such as Momentum, AdaGrad, Adam [7], learning rate schedules [8], or techniques such as batch normalization [9], online batch selection. Since the quality of a trained model depends strongly on the training data one of the most popular ways to regularize is regularization via data [2]. Data augmentation with traditional transformations such as shear, zoom in, rotation, hue or contrast modification, is widely used to increase a dataset size, even when the original set is large enough [10]. In the case of a class imbalance, a common regularization strategy is to undersample or oversample the data to improve generalization abilities [2]. Although their many advantages in some cases simple classical operations are not enough to significantly improve the neural network accuracy or to overcome the problem of overfitting. Moreover, they do not bring any new features to the images that could significantly improve the learning abilities of algorithm used as well as further generalization abilities of the networks [10].

A main goal of this paper is to propose an efficient regularization technique based on data augmentation, which uses the neural style transfer (NST) to synthesize rich in information and features images [11]. In order to improve the Convolutional Neural Network (CNN) generalization abilities, synthetized images are used to train the CNN instead of the real images, while real images are used only for validation and testing, what we think is an unconventional but interesting approach. As a case study, we choose very challenging benchmark dataset with skin lesions used for classification and segmentation tasks [12]. The main contributions of this paper are:

- The proposition of using neural style transfer as a training data augmentation method (skin lesion case study: benign skin lesion to malignant lesion);
- Incorporating unlabeled data into the training;

- Using only synthetized images as CNN's training set while using real images as a validation set;
- Proving that above mentioned significantly improved an accuracy as well as the repeatability of deep neural networks.

The reminder of the paper is as follows: in section II we summarize briefly a current state-of-the-art regarding the image augmentation methods. Later, we present a fresh look at the style-transfer method and a prove that it can be used as a data augmentation method (section III). Moreover, we present a methodology how to use neural style transfer method to generate new, unlabeled data and how to incorporate them into the training (section IV). In the last two sections, we present our results and bring a discussion about the advantages and drawbacks of our method.

## II. DATA AUGMENTATION METHODS

In general, the image synthesis methods that made a great impact in deep learning might be grouped into three categories: traditional transformations, advanced transformations and neural-based transformations.

### A. Traditional transformations

Traditional transformations are the most common data augmentation methods applied in deep learning. Traditional transformations are mainly defined as affine and geometric (elastic) transformations. The category of affine transformation methods covers techniques that use linear transformations to the image, such as rotation, shear, reflection, scaling [10,13], while the most popular geometric transformations are hue modification, contrast modification, white-balance, sharpening [14]. Both affine and elastic distortions or deformations are commonly used to increase the number of samples for training the deep neural models [15], to balance the size of datasets [16] or to increase the robustness of the model [14].

### B. Advanced transformations

The last years saw also more sophisticated transformations that are not as popular, commonly used and well-known as traditional transformations. We would like to present a few interesting approaches which can be used as data augmentation techniques. Goal of using hose method is to: make DNN's easier to train, with a greater ability to generalize, be more robust for input changes (what is important in a case of in adversarial attacks) [17].

One of the interesting approaches developed to make a CNN robust was data augmentation by adding a stochastic additive noise to an image [18]. Another approach is a random erasing technique proposed in [19], fast and relatively easy to implement yet giving good results in generalization ability of CNN's. In this method, one randomly paints a noise-filled rectangle in an image resulting in changing original pixels values. As the authors explained, expanding the dataset with images of various levels of occlusion reduces the risk of overfitting and makes the model more robust. Very surprising results were showed by authors in [20]. They create between-class images by mixing two images belonging to different classes with a random ratio and added them to a train set which significantly improved the network performance in the task of skin lesion classification.

### C. Neural-based transformations

One of the neural-based data augmentation approaches are the Generative Adversarial Networks (GANs) which are popular, yet a relatively new tool to perform unsupervised generation of the new images using min-max strategy [21]. GANs are found to be very useful in many different image generation and manipulation problems like text-to-image synthesis [22], super-resolution (generating high-resolution image out of low-resolution one) [23], image-to-image translation (e.g. convert sketches to images) [24], image blending (mixing selected parts of two images to get a new one) [25], image inpainting (restoring missing pieces of an image) [26] and unconditional [27] and conditional synthesis [28]. Recently, GANs started gaining attention in the field of medical image processing to solve tasks such as de-noising [29], segmentation [30], detection [31] and classification [32].

Another CNN-based method is an algorithm of artistic style based on a deep neural network that creates artistic images of high perceptual quality [11]. The first proposition of using CNN feature activations to mimic an artistic style of one image while maintaining the content of another was published in 2015 [11], and since then many different modifications to this method emerged such as: universal style transfer, semantic style transfer (ST), instance ST, doodle ST, stereoscopic ST, portrait ST, video ST and many more [33].

## III. DATA SYNTHESIS WITH A STYLE TRANSFER

### A. Fundamentals

The goal of the image style transfer is to mimic the style of one image while keeping the content of another one. Developed in the past, style transfer methods were mostly based on a texture transfer, where one copy the small fragments of one image to another [34]. Although in some cases it looks well and gives the impression of visually pleasing style transfer, it does not bring any new information to those pictures, but only copy and paste small patches. Such approach does not enrich the training data set with new information and therefore does not help in teaching the network. Since 2015, a new way of a style transfer named the neural style transfer started to gain popularity [11]. The idea was based on the Convolutional Neural Network's properties to carry the information about the data it was trained on within convolutional layers. First CNN's filters represent detailed pixel values, but deeper into the network, the more advanced and abstract features start to emerge. Therefore, neural style transfer combine properties from the last layers of the CNN, that carry information about the content, and from first layers that carry information about the style of an image. The method allows to recombine those features to achieve an image with a content of one photography but with a style of another (Fig. 1). To combine the content and the style of an image with NST, first one need to train CNN of a choice on a selected dataset (for example benchmark dataset like ImageNet).

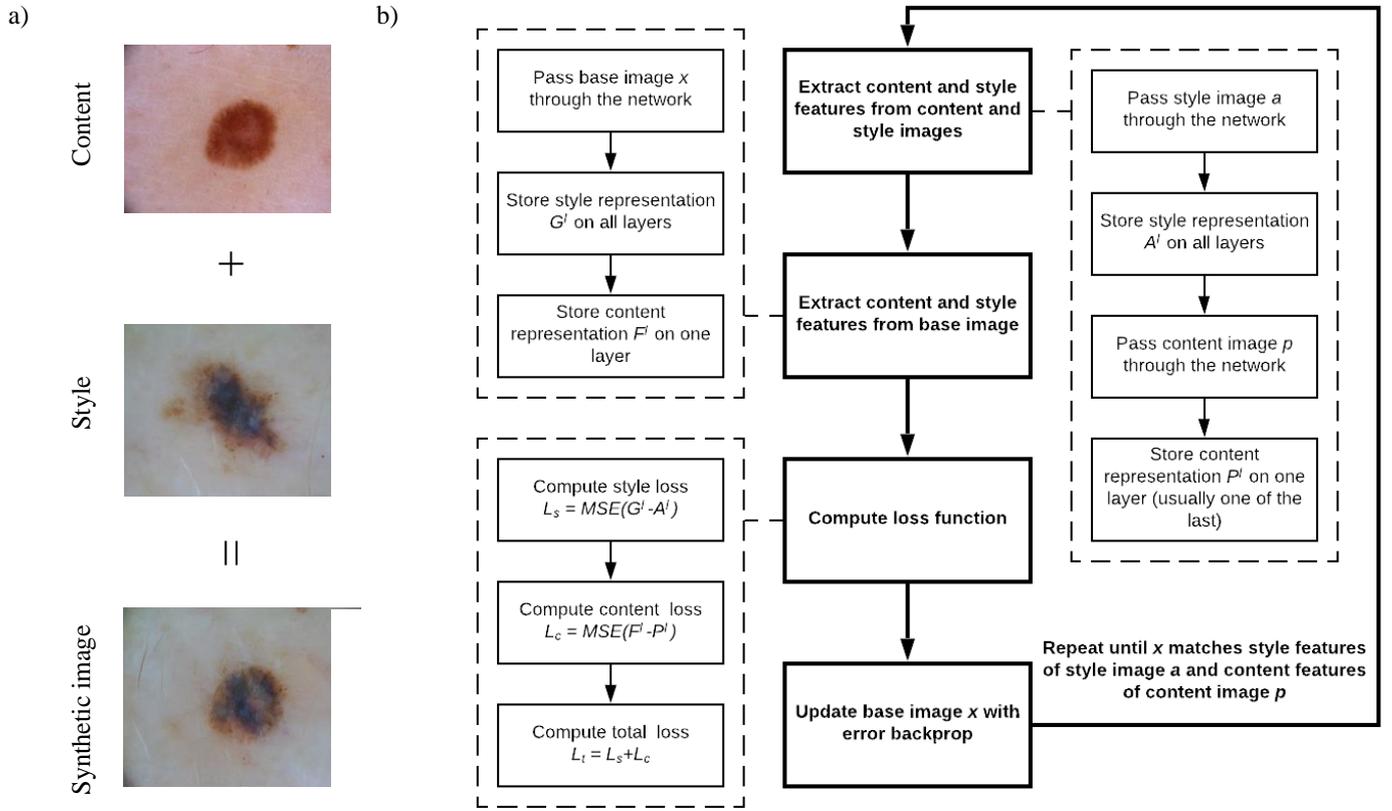

Fig. 1. a) Synthetic image created on the base of content and style images – Skin lesion example; b) Neural style transfer flowchart; MSE- Mean Square Error; $x$ – base image, $a$ – style image, $p$ – content image, $A^l$ - style representation; $P^l$- content representation; $G^l$ – style representation of a base image; $F^l$ – content representation of a base image; $L_s$ – style loss, $L_c$ – content loss, $L_t$ – total loss.

Then, both images of a style $a$ and a content $p$ should be passed through the CNN in order to calculate and save their filter responses. For the style representation $A^l$ usually all, or almost all layers are stored, while for the content representation $P^l$ only one of the last layers are saved. Next, in the same manner, the base image $x$ (which for instance white-noise image or content image) is passed through the network and its content $G^l$ and style $F^l$ representations are saved. Saved representations are later used to calculate a style and a content loss functions which are equal to mean square error of representations. The base image is updated with an error backpropagation, where the total loss $L_t$ function is equal to the sum of a content loss $L_s$ and a style loss $L_s$. The neural style transfer flowchart is presented in the Fig. 1b.

*B. Style transfer for data augmentation*

If both style and content images are chosen wisely, the style transfer allows to, for instance, change the illumination of one image, to create a retouch effect or to mimic the artistic style. Although NST gives many possibilities it is still underestimated in the field of data augmentation and is mainly used for artistic purposes. The flowchart of the neural style transfer method with the artificial image of skin lesion synthesis example is presented in Fig. 1.

Medical databases are often highly imbalanced in terms of a number of images in each class, for example in public skin lesion datasets there is usually about ten times less malignant lesions than benign lesions. In this paper, we propose to use neural style transfer to synthesize a skin lesion dataset. Instead of undersampling or oversampling the images to balance a number of images in each class what is a popular approach [35], we decided to synthesize it by merging the content of a benign lesion with a style of malignant one. It effected in the creation of a new image, with properties of both images (including possible artifacts such as dense hair, water bubbles, gel).

a)

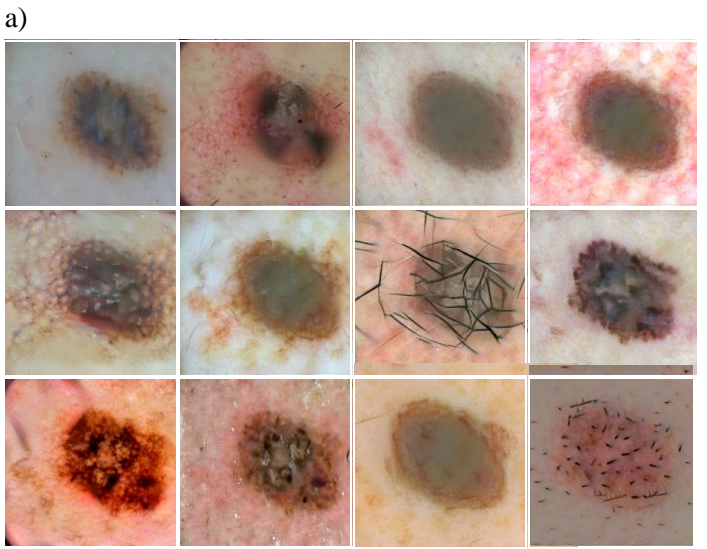

b)

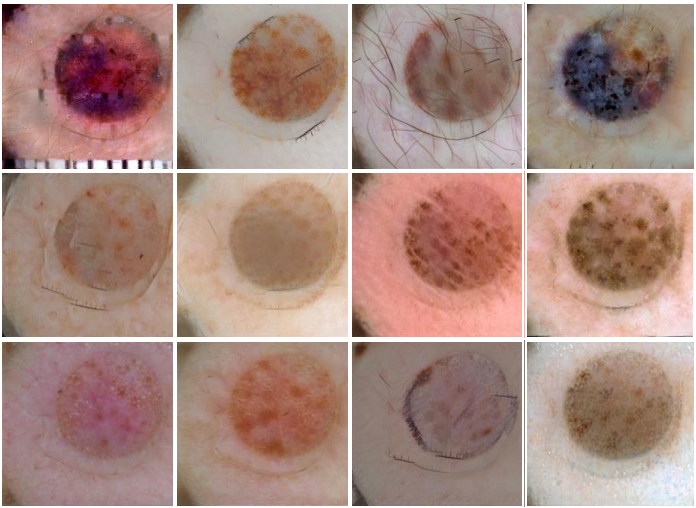

Fig. 2. The examples of synthetic images generated with the same content but with the different styles – a) first example, b) second example.

## IV. EFFICIENT METHOD OF NEURAL NETWORK TRAINING

### A. Idea of the methodology

The general idea of this method is to increase the amount of data further used to train the deep neural networks (data augmentation) but in the way to ensure an increase in the amount and variety of information that the generated images carry. It should lead to improvement of the images classification accuracy. The method consists of four subtasks: data augmentation, data labelling, dataset splitting and CNN's training (Fig. 3).

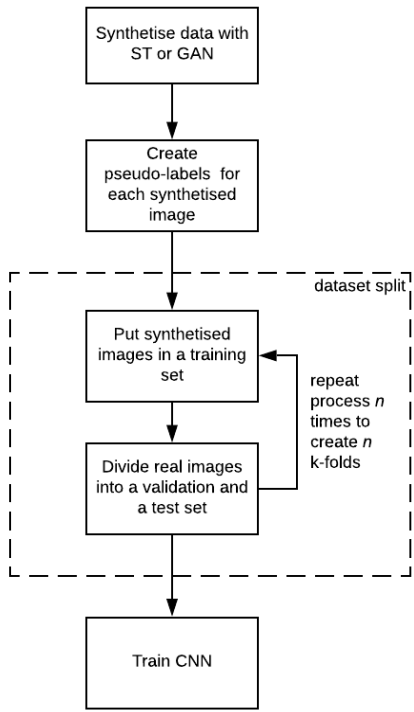

Fig. 3 The general idea of the method.

The first subtask is to generate new images by merging the style and content of two different images. In order to achieve this, the neural style transfer method [11] was employed [36]. This method does not simply copy-and-paste patches of an image but allows generating new information to the synthetized images. Moreover, enables for producing more visually pleasing results than traditional style transfer techniques Better insight into this phase was presented in section III. After synthetizing the data, we labelled them. The details are presented in section IV. Finally, we trained our CNN and saved the results. The proposed approach is a new regularization technique that can be successfully used among others and still improve the results. Common regularization techniques such as data augmentation based on affine transformations, dropout, and early stopping are exploited in all tests along with an artificial dataset [37].

### B. Data labelling

The downside of the style transfer data augmentation method is that generated images are unlabeled hence unusable in the supervised training. Our goal was to incorporate all of the unlabeled synthetized data into the training hence we decided to label images. We wanted to avoid manual labelling, so in order to do that we classified and labelled generated images with a DNN that was previously trained on the target dataset to properly classify the skin lesions. Moreover, we modified and selected a new DNN classification threshold in such way so the number of images in all classes are equal. It resulted in applying very noisy psuedo-labels, but the problem of uncertain labelling was solved by the dataset splitting described in the next section.

### C. Dataset splitting

We prepared the training, validation and test sets, using both real and generated data. All of the generated images were selected as a training set, while all real data was divided into validation and test sets. This type of data splitting is robust to uncertainty of labelling which exists in training set, because during the training DNN is validated only on the validation set (real annotated data). In total, we prepared 5-fold cross-validation [38] sub-sets. We paid special attention to avoid a situation where any images that were used to generate the augmented database did not appear in the validation or test sets (data leakage phenomena [39]).

### D. Implementation details

To synthesize the images we used the method of artistic neural style transfer based on [11] with improvements detailed in [40], implemented in Keras 2.0. In details, we used VGG16 pertained on the ImageNet, with the input image size of 224x224 px, Conv5_2 as the content layer. Each base image was initialized by the content picture, the content weight equal 0.025, style weight equal 1.0, and pooling layer with pool by the maximum operator. We aimed to draw more general conclusions and therefore we conducted training and tests on four different architecture types, namely: VGG9, VGG11,

VGG16 and DenseNet121 [41,42]. Each experiments were validated on five different testing folds.

## V. RESULTS

Through the method of neural style transfer we have augment the data set from 1088 malignant skin lesions and 12433 benign original images up to 248 489 synthetized one, while we used only 418 malignant lesion from the original dataset.

Next, we evaluated the performance of four popular and commonly used architecture types. We trained it on 54030 synthetized images, validated on 1976 real images, and tested it on 200 images per each of five folds.

We applied following regularization techniques to every tested network: traditional data augmentation (rotation, zoom, shear, reflection), dropout, early stopping. We gathered the results in Table I and in graphical form in Fig. 4.

TABLE I.  RESULTS OF EXPERIMENTS WITH AND WITHOUT DATA AUGMENTATION

| | Architecture | Fold number | | | | | |
|---|---|---|---|---|---|---|---|
| | | 0 | 1 | 2 | 3 | 4 | AVER |
| Without Data Augmentation | VGG9 | 0.805 | 0.820 | 0.845 | 0.845 | 0.805 | **0.824** |
| | VGG11 | 0.820 | 0.799 | 0.805 | 0.840 | 0.845 | **0.822** |
| | VGG16 | 0.830 | 0.850 | 0.815 | 0.875 | 0.840 | **0.842** |
| | DenseNet121 | 0.820 | 0.825 | 0.810 | 0.850 | 0.810 | **0.823** |
| With Data Augmentation | VGG9 | 0.835 | 0.880 | 0.845 | 0.875 | 0.850 | **0.857** |
| | VGG11 | 0.815 | 0.895 | 0.870 | 0.885 | 0.840 | **0.861** |
| | VGG16 | 0.825 | 0.850 | 0.835 | 0.830 | 0.815 | **0.831** |
| | DenseNet121 | 0.870 | 0.895 | 0.865 | 0.855 | 0.860 | **0.869** |

Networks trained with additional DA increased in area under the curve (AUC) on average by 2.68% comparing with networks trained traditionally. What is more, we observed that DNN's trained with DA tend to not overfit so quickly as in the case of the traditional approach. CNN trained without the additional DA increased its training accuracy after a few epochs, while validation's accuracy (ACC) dropped. In opposite, CNNs with DA slowly gained on both validation and training ACC what is desired situation during the training.

Less overfitting makes DNN generally more robust and they better generalize the new, unseen by DNN samples. Interestingly, the architecture VGG-16, which is bigger in terms of a number of parameters received slightly lower AUC than the approach without DA.

Moreover, smaller architectures such as VGG-9 and DenseNet trained with DA achieved better results than bigger architecture VGG-16 without DA.

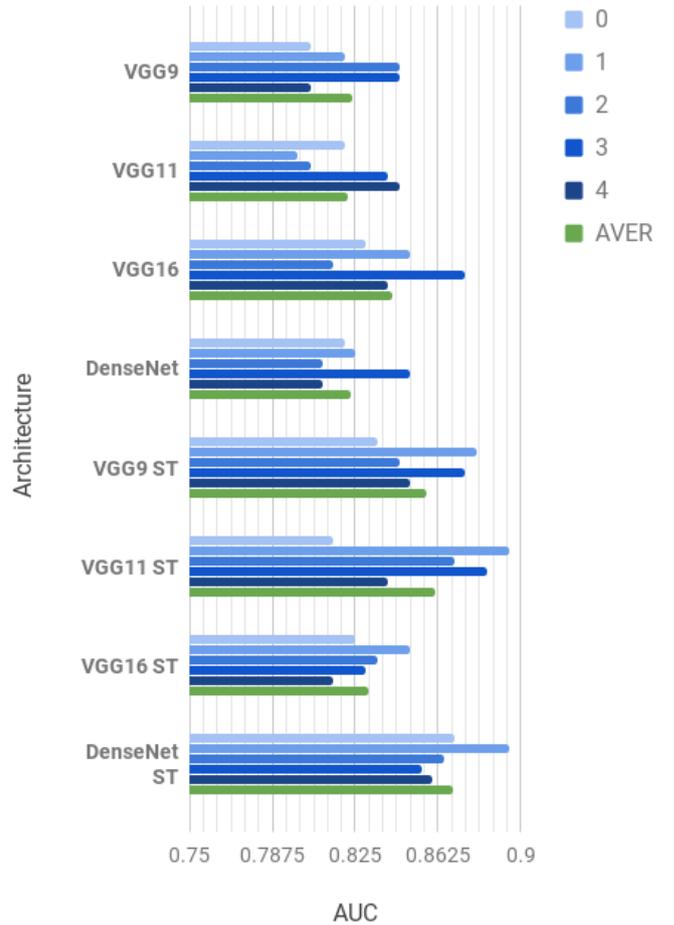

Fig. 4 AUC for evaluated architectures with or without style transfer (ST) for 5 k-folds

## VI. SUMMARY AND CONCLUDING REMARKS

In the paper we presented the current state-of-the-art of regularization techniques along with a variety of different data augmentation techniques.

We proposed our own regularization technique based on the data synthesis and we tested it on the very demanding case of skin lesion classification. We created the images with a neural style transfer where we mixed benign content and malignant style to synthetize new skin lesion images. Next, we created the pseudo-labels for each image by using CNN and prepared specific dataset split – we put all of the synthetized images into a training set, while the real images were split into validation and test set. At the end, such prepared data were used to train four different representative types of convolutional neural networks architectures. The experiments carried out for all architectures and for 5-folds have showed the average of 2.68% improvement of AUC comparing with the approach without proposed regularization technique.

Our NST regularization method allows to be used for image analysis, where NST gives satisfying results [11] and for other applications e.g. video style transfer for scene recognition, speech style transfer for speech recognizing or others. Moreover, our method instead of NST can be used with GANs [20] or other methods which allows to create additional unannotated data, or even where the new data is available but it is unlabeled or labeled with high uncertainty. NST regularization method should be used not instead but among other regularization techniques.

While this method is fast and easy to implement to any architecture, it is especially useful in case of using unpopular or specifically designed CNN architecture, where it is not possible to download pretrained CNN from the web, and using transfer learning is available only after additionally pretraining it. Hence, it is faster to train than pretraining it and then training it, it is especially useful to use it with the task of neural architecture search [43]. Also, when needed, it can be also used with the transfer learning.